\documentclass[letterpaper, 10 pt, conference]{ieeeconf}  

\IEEEoverridecommandlockouts                              
\usepackage{times}
\usepackage{multicol}
\usepackage{latexsym}
\usepackage{booktabs}
\usepackage{amsmath}
\usepackage{cite}     
\usepackage{listings}
\usepackage{mathrsfs}
\usepackage{amssymb}
\usepackage[ruled]{algorithm2e}
\usepackage{hyperref}
\usepackage[table]{xcolor}
\usepackage{graphicx}
\usepackage{amsthm}
\usepackage{subcaption}
\usepackage{siunitx}
\usepackage{amsfonts}
\usepackage[most]{tcolorbox}
\usepackage{float}

\theoremstyle{definition}
\newtheorem{definition}{Definition}

\theoremstyle{remark}
\newtheorem{remark}{Remark}

\theoremstyle{assumption}
\newtheorem{assumption}{Assumption}

\title{\LARGE \bf
See Something, Say Something: Context-Criticality-Aware Mobile Robot Communication for Hazard Mitigations}

\author{Bhavya Oza$^{2}$, Devam Shah$^{2}$, Ghanashyama Prabhu$^{2}$, Devika Kodi$^{2}$,
        \\and Aliasghar Arab$^{1,2*}$%
\thanks{This work was supported by the Department of Mechanical and Aerospace Engineering,
        New York University Tandon School of Engineering, and a Nokia Bell Labs Research Award.}%
\thanks{$^{1}$Aliasghar Arab is with The City College of New York, Grove School of Engineering,
        New York, NY 10031, USA.
        {\tt\small aarab@ccny.cuny.edu}}%
\thanks{$^{2}$Bhavya, Devam, Ghanashyama, Devika, and Aliasghar are with ASAS LABS,
        Department of Mechanical and Aerospace Engineering,
        Tandon School of Engineering, New York University,
        6 MetroTech Center, Brooklyn, NY 11201, USA.
        {\tt\small \{bpo7912, ds7855, ip2499, dak9250, aliasghar.arab\}@nyu.edu}}%
}
\begin{document}
\maketitle

\begin{abstract}
The proverb ``see something, say something'' captures a core responsibility of autonomous mobile robots in safety-critical situations: when they detect a hazard, they must communicate---and do so quickly. In emergency scenarios, delayed or miscalibrated responses directly increase the time to action and the risk of damage. We argue that a systematic context-sensitive assessment of the criticality level, time sensitivity, and feasibility of mitigation is necessary for AMRs to reduce time to action and respond effectively. This paper presents a framework in which VLM/LLM-based perception drives adaptive message generation, for example, a knife in a kitchen produces a calm acknowledgment; the same object in a corridor triggers an urgent coordinated alert. Validation in 60+ runs using a patrolling mobile robot not only empowers faster response, but also brings user trusts to 82\% compared to fixed-priority baselines, validating that structured criticality assessment improves both response speed and mitigation effectiveness.
\end{abstract}

\section{INTRODUCTION}

Mobile robots are increasingly deployed across public spaces, healthcare facilities, industrial sites, and emergency response environments\cite{fateh2015robust,ERMES2024,LangGroundSurveyIJCAI2024}---and their scale and autonomy will only grow. As they operate alongside people, AMRs have a unique opportunity to act as proactive safety agents: detecting risks before they escalate and coordinating mitigation with nearby individuals, responders, and infrastructure. The proverb ``see something, say something'' captures this responsibility precisely---but in safety-critical situations, \emph{how} and \emph{what} the robot says matters as much as whether it speaks at all. A knife in a kitchen and a knife in a corridor are the same object; the response must not be. Without a systematic framework to assess a situation's criticality---how dangerous it is, how urgent the response must be, and who can best act on it---robots default to fixed, context-free alerts that cause alarm fatigue, erode trust, and increase time to action\cite{SafetyTrustHRIReview2024,IntentCommWarehouse2024,campagna2025systematic}.

\begin{figure}[t!]
\centering
\includegraphics[width=1.0\linewidth]{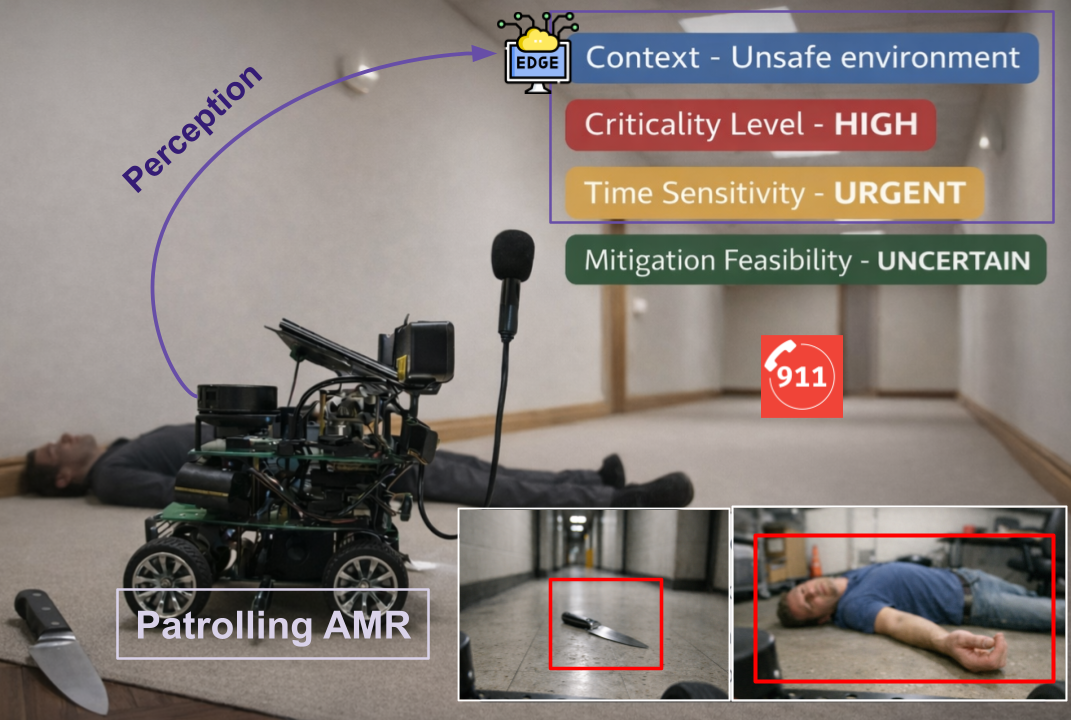}
\caption{A patrolling AMR detecting a fallen person, correlating a nearby knife with heightened risk, and triggering a context-aware mitigation response.}
\label{fig:social}\end{figure}

Crisis management research establishes that effective emergency response depends on four sequential stages: detection of risk, recognition and interpretation of that risk in context, communication of risk to the relevant parties, and mobilization of a coordinated collective response\cite{comfort2007crisis}. Critically, communication at the wrong register---too alarming for a low-risk event, or too muted for a high-risk one---induces unnecessary panic, erodes responder trust, and delays the mobilization of the right mitigation resources\cite{comfort2007crisis,branda2025role,PoliteEmergencyEvac2025}. Recent Vision-Language Models (VLMs) and Large Language Models (LLMs) make context-sensitive communication possible for embodied robots: VLMs interpret scenes by combining visual perception with semantic understanding\cite{BLIP2022,BLIP2_2023,PaLME2023}, while LLMs generate context-aware, human-readable guidance calibrated to assessed risk\cite{SayCan2022,RT2_2023}. Notably, these foundational models can distinguish a low-risk situation from a high-risk one involving the same object, going well beyond classical object-level approaches\cite{elhafsi2023semantic,ji2022proactive,kang2025real}. Yet existing robot communication systems remain largely rule-based and non-adaptive, lacking a principled mapping from context to mitigation action\cite{LanguageCommunicationSurvey2024,LLMRoboticsIntegrationSurvey2024,VLAChallenges2025}.

This paper proposes a systematic framework in which an AMR assesses each situation through three formally defined factors---criticality level, time sensitivity, and feasibility of mitigation---and uses foundational models to select the best mitigation response within its capabilities. Building on prior work on social navigation\cite{sotomi2025transparent,arab2025trust}, proactive safety\cite{maithani2026proactive}, and collaborative robot safety\cite{badguna2025virtual}, the framework adjusts message tone, character, and content to the computed criticality: a calm inquiry for low-risk situations, an urgent coordinated alert for high-risk ones. Implemented on a patrolling mobile robot with VLM-based scene understanding and LLM-based reasoning\cite{PaLME2023,RT2_2023}, validation across 60+ runs demonstrates 82\% user trust, a 10\% gain in detection accuracy, and strict compliance with escalation constraints\cite{SafetyTrustHRIReview2024,PromisesTrustHRI2021}---validating that systematic, context-sensitive criticality assessment empowers faster hazard mitigation and improves mitigation effectiveness.

\section{PROBLEM FORMULATION}
\label{sec:problem}

We formalize context-criticality-aware robot communication as a four-step sequential decision problem, grounded in the crisis management framework of \cite{comfort2007crisis}: (i) \emph{detect} a hazard from the robot's observation, (ii) \emph{recognize and interpret} the risk by extracting three contextual factors capturing how dangerous the situation is, how urgent the response must be, and who can best mitigate it, (iii) \emph{communicate} risk through a message whose tone and character are calibrated to avoid panic while conveying the correct level of urgency, and (iv) \emph{mobilize} the appropriate responders by routing the message to the right recipients. Section~III describes how foundational models realize each step.

\subsection{Step 1 — Hazard Detection}

Let $\mathcal{O}_t$ denote the robot's multimodal observation at time $t$ (RGB image, visual attention map, spatial context). Perception maps this observation to a hazard label:

\begin{definition}[Hazard Detection]
\begin{equation}
a_t = f_{\text{hazard}}(\mathcal{O}_t), \quad a_t \in \mathcal{A}
\label{eq:hazard_detection}
\end{equation}
where $\mathcal{A}$ is the discrete set of hazard categories (e.g., sharp object, person down, unattended item).
\end{definition}

\begin{remark}[Detection is Not Enough]
Knowing \emph{what} was detected ($a_t$) is necessary but not sufficient to determine the right response. The same object detected in different environments requires a fundamentally different communication action. The three contextual factors below resolve this ambiguity.
\end{remark}

\subsection{Step 2 — Contextual Factors}

\begin{definition}[Contextual Factors]
Given $a_t$ and $\mathcal{O}_t$, context understanding extracts the triple:
\begin{equation}
\theta_t = (d_t,\, \tau_t,\, \phi_t) \in \Theta
\label{eq:context_factors}
\end{equation}
The three factors are defined as follows.

\textbf{Criticality Level} $d_t \in \{\text{Low},\, \text{Medium},\, \text{High}\}$ reflects the inherent severity of the detected hazard within its environmental context. The same object may carry a different severity depending on location and surrounding activity.

\textbf{Time Sensitivity} $\tau_t \in \{\text{Immediate},\, \text{Soon},\, \text{Near Future}\}$ captures urgency. \textit{Immediate} indicates imminent harm requiring instant action; \textit{Soon} denotes a risk likely to materialize within a short window; \textit{Near Future} indicates a potential risk that is not yet pressing but should be logged.

\textbf{Feasibility of Mitigation} $\phi_t \in \{\text{Robot},\, \text{PoC},\, \text{Help Needed}\}$ identifies who is best positioned to resolve the hazard: the robot itself, a nearby Person of Contact (PoC), or external emergency responders when the situation exceeds local capacity.
\end{definition}

\subsection{Step 3 — Risk Assessment}

The three contextual factors, together with broader environmental context $\theta_{\text{env}}$ (location type, crowd density, presence of vulnerable individuals), determine the overall criticality:

\begin{definition}[Risk of Loss Criticality]
\begin{equation}
k_t = f_{\text{criticality}}(d_t,\, \tau_t,\, \phi_t,\, \theta_{\text{env}})
\label{eq:criticality}
\end{equation}
\end{definition}

\begin{remark}[Context-Dependence of Criticality]
The same physical hazard $a_t$ yields a different criticality $k_t$ depending on the context; for example, a knife during the preparation of food in a kitchen yields $d_t = \text{Low}$, $\tau_t = \text{Near Future}$, $\phi_t = \text{Robot}$, and thus $k_t = \text{Low}$; the same knife in a public corridor yields $d_t = \text{High}$, $\tau_t = \text{Immediate}$, $\phi_t = \text{Help Needed}$, and thus $k_t = \text{High}$. Section~III shows how an LLM realizes $f_{\text{criticality}}$ from scene context.
\end{remark}

\subsection{Step 4 — Criticality-Aware Communication}
\begin{definition}[Communication Tuple]
A robot-to-human message at time $t$ is a structured tuple:
\begin{equation}
\mathcal{M}_t = (m_t,\, \gamma_t,\, \chi_t) \in \mathcal{C}
\label{eq:message_tuple}
\end{equation}
where $m_t$ is the \textbf{message text} (clear, precise, and instructive), $\gamma_t \in [0,10]$ is the \textbf{tone} (communication intensity, scaled by $k_t$), and $\chi_t \in \{\text{inquiry},\, \text{alert},\, \text{urgent}\}$ is the \textbf{character type} (matched to $k_t$). Section~III specifies the exact mapping from $k_t$ to each component.
\end{definition}

The complete communication output pairs the message tuple with a recipient set $r_t \subset \{\text{nearby},\, \text{remote},\, \text{coordination}\}$, also determined by $k_t$:
\begin{equation}
c_t = (\mathcal{M}_t,\, r_t)
\label{eq:comm_output}
\end{equation}

\subsection{Objective}

The robot selects a communication policy $\pi$ over the horizon $[0,T]$ to minimize expected hazard loss while penalizing alarm fatigue:
\begin{equation}
\min_{\pi}\; \mathbb{E}\!\left[\sum_{t=0}^{T}
\Bigl(L_{\text{hazard}}(\mathbf{S}_t, c_t)
+ \lambda\, L_{\text{fatigue}}(c_{0:t})\Bigr)\right]
\label{eq:objective}
\end{equation}
where $\mathbf{S}_t = (a_t, \theta_t, k_t)$ is the system state at time $t$, $L_{\text{hazard}}$ captures residual risk or mitigation failure when the message fails to prompt timely action, $L_{\text{fatigue}}$ penalizes excessive or repeated alerts that erode human trust, and $\lambda > 0$ balances safety urgency against communication fatigue.

\begin{assumption}[Communication Effectiveness]
Policy performance is quantified by a scalar $\varepsilon \in [0,1]$ defined as the weighted combination of four sub-metrics:
\begin{equation}
\varepsilon = w_1\,\varepsilon_{\text{det}} + w_2\,\varepsilon_{\text{msg}} + w_3\,\varepsilon_{\text{coord}} + w_4\,\varepsilon_{\text{lat}}, \quad w_i = 0.25\;\forall i
\label{eq:effectiveness}
\end{equation}
where $\varepsilon_{\text{det}}$ is detection accuracy, $\varepsilon_{\text{msg}}$ is criticality-message alignment, $\varepsilon_{\text{coord}}$ is coordination success rate, and $\varepsilon_{\text{lat}} = 1 - T_{\text{total}}/T_{\text{max}}$ is latency compliance. Equal weights ($w_i = 0.25$) are used in the current implementation, treating each sub-metric as equally important; domain-specific weighting is a direction for future work. The system is inactive when $\varepsilon = 0$. A higher $\varepsilon$ indicates better alignment between criticality $k_t$ and communication impact. Section~IV reports $\varepsilon$ as the primary evaluation metric.
\end{assumption}

\section{METHODOLOGY}
\label{sec:methodology}
The objective is to realize the mapping $(a_t, \theta_t) \mapsto (k_t, c_t)$ via foundational models while enforcing the alarm constraint Eq.~(\ref{eq:alarm_constraint}) and maximizing $\varepsilon$ Eq.~(\ref{eq:effectiveness}). The system consists of five components: (1) multimodal hazard detection; (2) context-aware risk assessment; (3) criticality-driven message generation; (4) multi-party communication; and (5) integration with the autonomous navigation stack.

\subsection{Hazard Detection Model}

Let $X \in \mathbb{R}^{m \times n \times 3}$ denote the raw image. A heatmap $H = g(X)$ highlights salient regions via Grad-CAM, and a scene caption $L = f(X, H)$ is generated by Bootstrapped Language-Image Pre-training (BLIP). The LLM fuses both to infer the hazard:
\begin{equation}
a_t = \text{LLM}(\psi(H, X)), \quad a_t \in \mathcal{A}
\label{eq:llm_hazard}
\end{equation}
where $\psi(H,X)$ fuses visual saliency and scene context.

The same LLM call also extracts the contextual factors $\theta_t = (d_t, \tau_t, \phi_t)$ as structured outputs: the prompt (Section~\ref{sec:experiments}) instructs the model to jointly infer hazard category, criticality level $d_t$, time sensitivity $\tau_t$, and feasibility of mitigation $\phi_t$ from the fused visual-linguistic context. This realizes Step~2 of the Problem Formulation (Section~\ref{sec:problem}) within a single inference pass.

\begin{remark}[Context Disambiguation]
By conditioning on both the visual heatmap $H$ and the scene caption $L$ (Eq.~(\ref{eq:llm_hazard})), the LLM can disambiguate objects that are superficially identical but contextually different---e.g., a knife in a kitchen versus a knife in a corridor---yielding different $a_t$ classifications and different $\theta_t$ extractions, which in turn produce different $k_t$ values via Eq.~(\ref{eq:criticality}).
\end{remark}

\subsection{Criticality-Driven Message Generation}

Given $a_t$ and $\theta_t = (d_t, \tau_t, \phi_t)$ (Section~II), the LLM returns a continuous risk score $\rho_t \in [0,10]$, which is thresholded to obtain the discrete criticality $k_t$:
\begin{equation}
k_t =
\begin{cases}
\text{Low},    & \rho_t \in [0,4]  \\
\text{Medium}, & \rho_t \in [5,7]  \\
\text{High},   & \rho_t \in [8,10]
\end{cases}
\label{eq:criticality_threshold}
\end{equation}
realizing $f_{\text{criticality}}$ from Eq.~(\ref{eq:criticality}). The three-band thresholds $[0$--$4]$, $[5$--$7]$, $[8$--$10]$ are consistent with standard three-tier risk severity classifications used in occupational safety frameworks and were empirically validated across the five experimental scenarios. Tone intensity is set to $\gamma_t = \rho_t$, directly coupling communication urgency to the numerical risk score.

\begin{remark}[LLM Consistency and Opacity]
Because $f_{\text{criticality}}$ is realized by a prompted LLM, the mapping is not analytically inspectable. Across repeated runs on identical inputs, the system produced consistent $k_t$ classifications; however, determinism is not guaranteed across API versions. This is a known limitation: the formal framework provides structural guarantees (alarm constraint Eq.~(\ref{eq:alarm_constraint}), recipient routing Eq.~(\ref{eq:recipients})), while LLM reliability remains an empirical property. Ablation studies isolating the individual contribution of $d_t$, $\tau_t$, and $\phi_t$ to $\rho_t$, and formal analysis of edge-case failure modes, are planned as future work.
\end{remark} The message text is generated by the LLM conditioned on the detected hazard, criticality, and environment:
\begin{equation}
m_t = \text{LLM}_{\text{comm}}(a_t,\, k_t,\, \theta_{\text{env}})
\label{eq:msg_generation}
\end{equation}

The tone $\gamma_t$ and character $\chi_t$  are then set according to $k_t$ by 
\begin{equation}
\gamma_t =
\begin{cases}
[0,4],   & k_t = \text{Low}    \\
[5,7],   & k_t = \text{Medium} \\
[8,10],  & k_t = \text{High}
\end{cases}
\label{eq:tone_character}
\end{equation}
and
\begin{equation}
\chi_t =
\begin{cases}
\text{inquiry}, & k_t = \text{Low}    \\
\text{alert},   & k_t = \text{Medium} \\
\text{urgent},  & k_t = \text{High}
\end{cases}
\label{eq:tone_character2}
\end{equation}

\noindent yielding the communication tuple $\mathcal{M}_t$ Eq.~(\ref{eq:message_tuple}). The audible alarm signal $a_t^{\text{alarm}} \in \{0,1\}$ is activated only for non-Low criticality:
\begin{equation}
a_t^{\text{alarm}} =
\begin{cases}
1, & k_t \in \{\text{Medium},\, \text{High}\} \\
0, & k_t = \text{Low}
\end{cases}
\label{eq:alarm_constraint}
\end{equation}
preventing unnecessary escalation and alarm fatigue. Recipient selection follows criticality:
\begin{equation}
r_t =
\begin{cases}
\{\text{nearby}\},                             & k_t = \text{Low}    \\
\{\text{nearby, remote}\},                     & k_t = \text{Medium} \\
\{\text{nearby, remote, coordination}\},       & k_t = \text{High}
\end{cases}
\label{eq:recipients}
\end{equation}

\begin{algorithm}[h!]
\caption{Context-Aware Criticality and Alarm-Constrained Communication}
\label{alg:CriticalityComm}
\nl Initialize VLM, LLM, Context Module, Communication Module\;
\nl Initialize alarm state $a_t^{\text{alarm}} \leftarrow 0$\;

\While{robot navigating}{
    \nl Capture observation $\mathcal{O}_t = (X,H)$\;
    \nl Detect hazard $a_t$\;

    \If{$a_t$ detected}{
        \nl Extract $\theta_t = (d_t, \tau_t, \phi_t)$\;
        \nl Compute criticality $k_t$ via Eq.~(\ref{eq:criticality})\;
        \nl Generate $m_t$, set $\gamma_t$, $\chi_t$, $a_t^{\text{alarm}}$, $r_t$\;
        \nl Deliver $c_t = (\mathcal{M}_t, r_t)$\;
        \nl Update navigation policy to avoid hazard\;
    }
    \Else{
        \nl Reset $a_t^{\text{alarm}} \leftarrow 0$\tcp*{clear alarm when no hazard present}
    }

    \nl Execute navigation step\;
}
\end{algorithm}

\subsection{Latency and Time-to-Communication}

Total communication latency is:
\begin{equation}
T_{\text{total}} = T_{\text{camera}} + T_{\text{heatmap}} + T_{\text{LLM}} + T_{\text{comm}} \leq T_{\text{max}}
\label{eq:latency}
\end{equation}
Higher-criticality messages are prioritized in scheduling to minimize $\sum_i T_i$ and maximize mitigation effectiveness; this priority scheduling directly reduces the time to first alert for urgent events compared to FIFO processing.
In our experimental implementation, $T_{\text{total}}$ (Eq.~(\ref{eq:latency})) averaged approximately 12 seconds with the BLIP + ChatGPT-4.0 pipeline. Onboard tasks ($T_{\text{camera}}$, $T_{\text{heatmap}}$ via Grad-CAM) completed within ${\sim}2.5$ s on the Jetson Nano; the remaining ${\sim}9.5$ s were attributed to cloud LLM/VLM API round-trips ($T_{\text{LLM}}$). The observed worst-case latency across all 60+ runs was below $T_{\text{max}} = 20$ s. To manage API variability, a fallback rule-based policy activates when $T_{\text{total}} > T_{\text{max}}$: the last known $k_t$ classification is used to dispatch a conservative pre-formulated alert, ensuring communication is never fully blocked by cloud latency.

\textbf{Deployment Tradeoff.} Table~\ref{tab:llm_comparison} shows that Gemini 2.0 Flash achieves ${\sim}5$ s latency, which may be preferable for High-criticality scenarios (e.g., person down) where the ${\sim}12$ s GPT-4.0 pipeline is safety-marginal. However, Gemini 2.0 Flash exhibited weaker visual-grounded reasoning in ambiguous cases (e.g., toy vs.\ real gun), where the BLIP + GPT-4.0 pipeline outperformed. A hybrid policy---using Gemini for initial fast triage and GPT-4.0 for disambiguation---is a promising direction for future work.

\subsection{System Implementation and Hardware Architecture}

The proposed context-aware framework is realized through a modular hardware-software stack designed for real-time edge-to-cloud processing, building upon established mobile robot control and safety standards.

\textbf{Hardware Configuration:} The system is deployed on a custom patrolling mobile robot centered around an \textit{NVIDIA Jetson Nano} compute module. Primary sensing is provided by an \textit{Intel RealSense RGB-D camera} for visual perception and a \textit{LiDAR} for spatial mapping and navigation . Human-robot interaction is enabled via a high-gain microphone and a localized speaker system to facilitate proactive intent communication.

\textbf{Software Pipeline:} Operating on \textit{ROS 1 Melodic} (note: ROS 1 reached end-of-life in May 2025; migration to ROS 2 is planned for future deployment), the software architecture consists of four primary layers:
\begin{itemize}
    \item \textit{Processing Pipeline:} Manages raw sensor streams to produce the multimodal observations $\mathcal{O}_{t}$ and visual attention maps $H$ using localized gradient-based methods.
    \item \textit{Cloud Integration Layer:} Acts as a secure gateway to route processed image features and captions to remote model endpoints, ensuring efficient management of computational loads.
    \item \textit{LLM/VLM Processing Engine:} Employs BLIP-base \cite{BLIP2022} (no task-specific fine-tuning) for visual-language grounding and the ChatGPT-4.0 API (gpt-4-0613 snapshot) for high-level reasoning and risk assessment. The system depends on the commercial OpenAI API, which is subject to version changes and rate limits; reproducibility is mitigated by pinning the model snapshot and providing the full prompt in supplementary material.
    \item \textit{Response Generation:} A prompt engineering layer formats the resulting criticality $k_t$ into the communication tuple $\mathcal{M}_t$. The final output is converted to speech to improve user understanding and assistance during potential hazards.
\end{itemize}
\subsection{Multi-Party Coordination}
\begin{figure*}[ht!]
    \centering
    \includegraphics[width=1\linewidth]{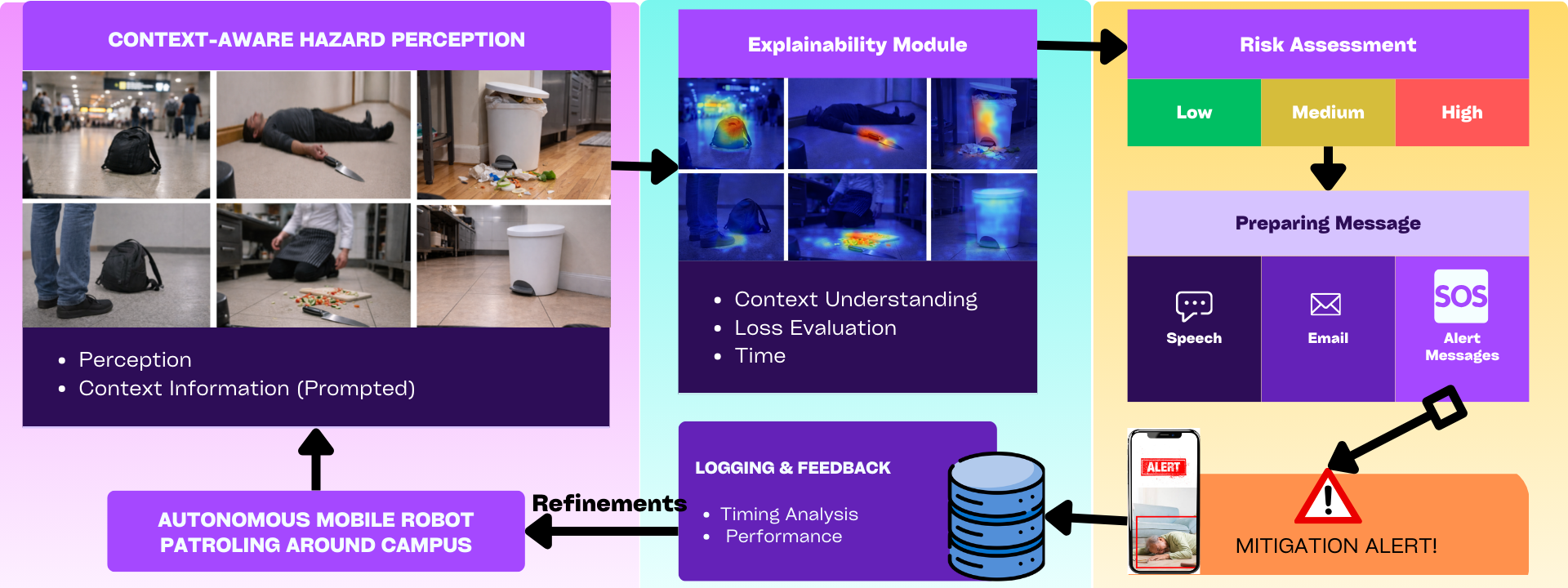}
    \caption{Workflow of hazard detection, communication strategy generation, and multi-party coordination.}
    \label{fig:workflow}
\end{figure*}

Communication is transmitted across three channels based on $r_t$: speech and visual alerts for nearby individuals, network messaging for remote responders, and API calls to the system for maintenance coordination. The complete pipeline is illustrated in Fig.~\ref{fig:workflow}.

\section{EXPERIMENTS AND RESULTS}
\label{sec:experiments}
We validate the core claim: \emph{when a robot sees something, what and how it says it must be adjusted to the criticality level $d_t$, time sensitivity $\tau_t$, and feasibility of mitigation $\phi_t$ to have maximum impact on hazard mitigation.} Experiments were conducted on a patrolling mobile robot running Robot Operating System (ROS) 1 Melodic with multimodal perception and LLM-driven communication, over 60+ runs in building corridors and a kitchen setting. The robot detects hazards via Eq.~(\ref{eq:hazard_detection}), extracts contextual factors Eq.~(\ref{eq:context_factors}), computes criticality Eq.~(\ref{eq:criticality}), and delivers a structured communication output Eq.~(\ref{eq:comm_output}).

\textbf{Baseline.} An object-level baseline uses the same perception pipeline but assigns a fixed criticality by object identity alone (e.g., ``knife detected'' $\rightarrow$ High, regardless of context), without LLM-based context reasoning. The complete pipeline reasons over the context of the scene to compute $\theta_t$ and $k_t$, producing differentiated responses to the same object. We acknowledge that this baseline is intentionally simple; a stronger rule-based contextual baseline using predefined location-to-criticality mappings (e.g., kitchen $\rightarrow$ Low, corridor $\rightarrow$ High) without LLM reasoning would better isolate the contribution of semantic understanding, and is planned for future evaluation.

\textbf{Scope.} The five scenarios cover a controlled but representative slice of indoor single-hazard situations. Simultaneous or cascading hazards, crowded multi-agent environments, and temporally escalating risks are not evaluated in this work and represent important directions for future empirical expansion.

\begin{table}[h]
\centering
\caption{Comparison of LLM Communication Models}
\label{tab:llm_comparison}
\scriptsize
\resizebox{\columnwidth}{!}{%
\begin{tabular}{|l|c|c|c|}
\hline
\textbf{Evaluation Criteria} & \textbf{Gemini 2.0 Flash} & \textbf{BLIP + ChatGPT-3.5} & \textbf{BLIP + ChatGPT-4.0} \\ \hline
Tone & Direct, fast & Balanced, adaptive & Calm, safety-oriented \\ \hline
Structure & Concise & Well-structured & Explicit hazard reasoning \\ \hline
Clarity & High (simple tasks) & Very high & Scenario-specific clarity \\ \hline
Context Handling & Moderate continuity & Strong reasoning & Visual-grounded reasoning \\ \hline
Latency & $\sim$5 s & $\sim$14 s & $\sim$12 s \\ \hline
\end{tabular}%
}
\end{table}
\begin{tcolorbox}[colback=gray!10!white, colframe=black, title=LLM Hazard Detection and Communication Prompt, sharp corners=south, boxrule=0.5 mm]
\textit{
"You are an indoor patrolling mobile robot performing context-aware hazard assessment. 
Given an image caption and a visual attention summary, infer the hazard category, contextual risk level, and appropriate mitigation action.
Estimate a risk score (0--10) based on environmental context and human activity.
Generate a decision containing hazard type, risk assessment, time sensitivity, feasibility of intervention, and a concise human-readable instruction.
Risk evaluation must be context-sensitive: the same object may be low risk in a safe context and high risk when associated with unsafe behavior or vulnerable individuals."
}
\end{tcolorbox}

\subsection{Scenarios: Context Determines What Gets Said}

Five scenarios directly test whether the three risk factors---$d_t$, $\tau_t$, $\phi_t$---correctly determine the message content and delivery channel (Table~\ref{tab:context_communication_strategies}):

\begin{itemize}
    \item \textbf{S1 -- Knife in unsafe area (High):} High $d_t$, Immediate $\tau_t$, Help Needed $\phi_t$ $\Rightarrow$ $k_t = \text{High}$, urgent alert, alarm activation, email to authorities.
    \item \textbf{S2 -- Knife in kitchen (Low):} Low $d_t$, Near Future $\tau_t$, Robot $\phi_t$ $\Rightarrow$ $k_t = \text{Low}$, inquiry message, no alarm.
    \item \textbf{S3 -- Person down (High):} High $d_t$, Immediate $\tau_t$, Help Needed $\phi_t$ $\Rightarrow$ $k_t = \text{High}$, urgent alert, remote responder notification.
    \item \textbf{S4 -- Toy / Fake gun (Low):} LLM disambiguates context $\Rightarrow$ $k_t = \text{Low}$, cautionary alert, no alarm escalation.
    \item \textbf{S5 -- Cleaning / Trash (Low):} Low $d_t$ $\Rightarrow$ $k_t = \text{Low}$, maintenance notification only.
\end{itemize}

\begin{table*}[t]
\centering
\caption{Context- and Criticality-Aware Communication Strategies}
\label{tab:context_communication_strategies}
\small
\begin{tabular}{|p{3.2cm}|p{5.5cm}|p{1.5cm}|p{5.9cm}|}
\hline
\textbf{Scenario} & \textbf{Context Interpretation} & \textbf{Risk Level} & \textbf{Communication Strategy} \\ \hline

Knife in Unsafe Surrounding 
& Sharp object detected near a person in a public or restricted area 
& High 
& Immediate voice alert, alarm activation, and automated email notification to authorities \\ \hline

Knife in Kitchen (Proper Use) 
& Knife identified in a designated kitchen area during food preparation 
& Low 
& No alarm triggered; contextual acknowledgment with optional safety reminder \\ \hline

Person Lying Down 
& Human detected in abnormal posture on the floor, possibly requiring assistance 
& High 
& Voice alert to nearby individuals and emergency notification with location details \\ \hline

People in Distress 
& Signs of panic or abnormal human activity detected in the environment 
& High 
& Clear assistance instructions via speech and urgent responder alert with contextual summary \\ \hline

Trash / Cleaning Activity 
& Garbage detected without immediate threat or hazardous context 
& Low 
& No alarm; maintenance notification sent as a non-urgent message \\ \hline

\end{tabular}
\end{table*}

\subsection{Context Improves What the Robot Detects}

Compared to the object-level baseline (70\% detection accuracy), the context-aware pipeline achieved \textbf{80\% accuracy} (+10\%) by reasoning about $d_t$, $\tau_t$, and $\phi_t$ rather than object identity alone. 

Among the evaluated models detailed in Table~\ref{tab:llm_comparison}, the BLIP + ChatGPT-4.0 pipeline was selected for deployment. Although ChatGPT-4.0 demonstrated strong linguistic performance, the BLIP-integrated approach provided more reliable vision-grounded reasoning and structured hazard-aware explanations. This multimodal reasoning capability resulted in improved contextual disambiguation and safer communication outputs in ambiguous scenarios. This improvement is most evident in visually ambiguous cases (e.g., toy gun vs.\ real gun), where contextual grounding Eq.~(\ref{eq:criticality}) prevents false escalations and reduces misclassifications.

\begin{figure}[h]
    \centering
    \includegraphics[width=1\linewidth]{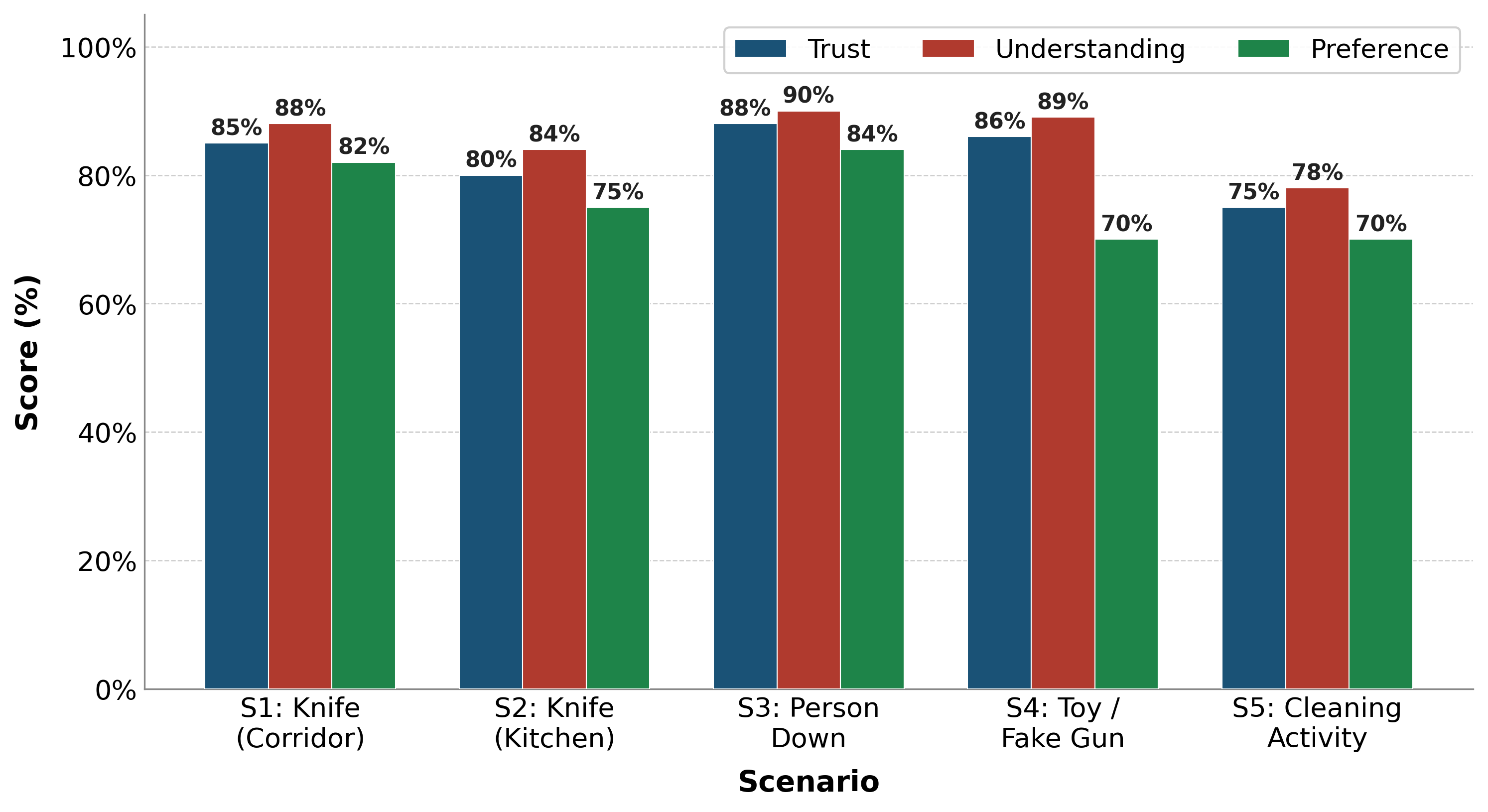}
    \caption{User Evaluation Breakdown (Trust, Understanding, Preference) for different scenarios}
    \label{fig:user_evaluation}
\end{figure}

\subsection{Context Determines How the Robot Responds}
\begin{figure}[h]
    \centering
    \includegraphics[width=1\linewidth]{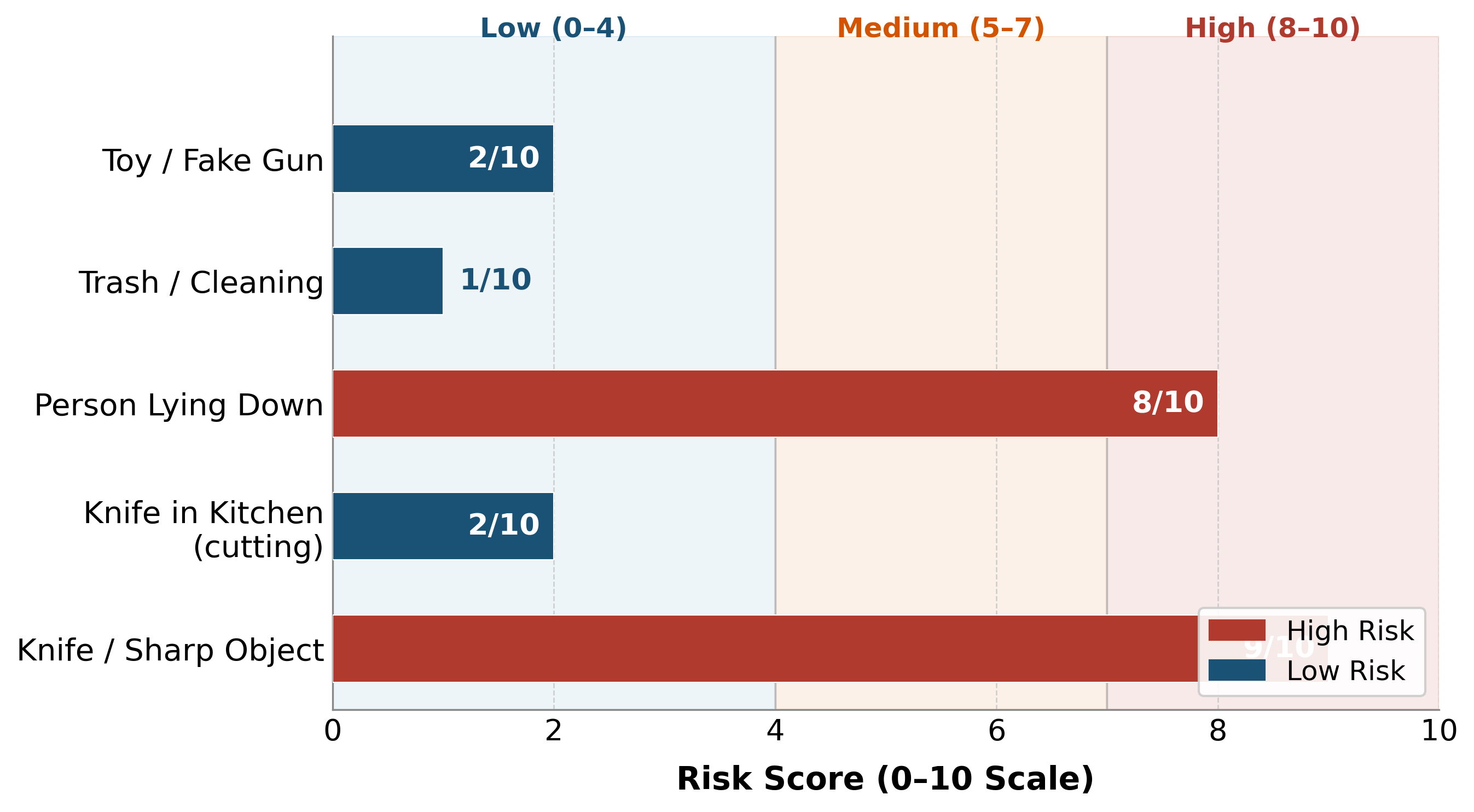}
    \caption{Scenario-wise risk scores $\rho_t$ (Eq.~(\ref{eq:criticality_threshold})): Low $[0\text{--}4]$, Medium $[5\text{--}7]$, High $[8\text{--}10]$. Alarm activation occurs only for $\rho_t \geq 5$.}
    \label{fig:risk_classification}
\end{figure}

Risk scores $\rho_t \in [0,10]$ (Eq.~(\ref{eq:criticality_threshold})) were correctly banded across all scenarios (Fig.~\ref{fig:risk_classification}), and the alarm constraint Eq.~(\ref{eq:alarm_constraint}) was strictly satisfied in every run---alarms activated only for $\rho_t \geq 5$ (i.e., $k_t \in \{\text{Medium, High}\}$), preventing unnecessary escalation in Low-criticality scenarios such as kitchen activity or garbage detection. Message generation Eq.~(\ref{eq:msg_generation}) and tone/character assignment Eq.~(\ref{eq:tone_character}) matched expected values in \textbf{82\% of cases}: High-criticality events produced urgent, high-intensity guidance; Low-criticality events produced calm, inquiry-level responses. Recipient selection Eq.~(\ref{eq:recipients}) correctly routed alerts in all runs. This confirms that $\gamma_t$ and $\chi_t$ effectively operationalize the criticality function and that context changes not just what is said, but how it is said.

\subsection{Human Impact: Trust, Understanding, and Effectiveness}

\textbf{Study Protocol.} The evaluation employed a within-subjects design in which participants were recruited from a university environment with varying levels of prior robot exposure (ranging from none to regular interaction). Each participant experienced both the proposed context-aware system and the fixed-priority baseline across four representative scenarios. Trust was assessed using 7-point Likert items adapted from the Jian et al.\ trust-in-automation scale; command understanding was measured via post-scenario comprehension questions; and preference was elicited through a direct forced-choice comparative rating after paired exposure. Participants acted as bystander observers and were not required to take physical mitigation action, consistent with the target deployment role of the robot as a proactive communicator in public environments.

User evaluation through structured sessions (Fig.~\ref{fig:user_evaluation}) yielded \textbf{82\% trust}, \textbf{85\% command understanding}, and \textbf{78\% preference} over fixed-priority systems. Compared to a static-priority baseline, the framework reduced unnecessary alarms and improved classification robustness by 10\%. The aggregated communication effectiveness factor $\varepsilon = 0.81$ reflects combined performance in detection accuracy, criticality correctness, message appropriateness, coordination success, and latency. Together, these results address both terms of the objective Eq.~(\ref{eq:objective}): the 10\% accuracy gain and 82\% criticality-message alignment reduce $L_{\text{hazard}}$ by ensuring timely, appropriate mitigation; strict alarm constraint compliance and reduced unnecessary escalation reduce $L_{\text{fatigue}}$, directly preserving human trust.

\section{CONCLUSIONS}
\label{sec:conclusion}

This paper formalizes context-criticality-aware communication for hazard mitigation as a sequential decision problem and implements it on a patrolling mobile robot using VLM/LLM-based foundational models. Without a principled mapping from context to communication, robots either over-escalate (causing alarm fatigue) or under-escalate (allowing preventable damage); the proposed framework avoids both failure modes through systematic assessment of criticality level $d_t$, time sensitivity $\tau_t$, and feasibility of mitigation $\phi_t$. Validation across 60+ runs yields $\varepsilon = 0.81$, 82\% user trust, a 10\% accuracy gain over fixed-priority baselines, and strict alarm constraint compliance in every run---confirming that context-sensitive criticality assessment empowers faster hazard mitigation. Future work will pursue factor ablation, broader scenario evaluation, formal verification of $f_{\text{criticality}}$, Partially Observable Markov Decision Process (POMDP)-based policies, and migration to ROS 2.
\section*{Acknowledgments}
The authors thank Nokia Bell Labs and Mike Coss for their invaluable feedback. The authors appreciate the support and encouragement of our colleagues in the Department of Mechanical \& Aerospace Engineering at New York University.

\bibliographystyle{unsrt}
\bibliography{references}

@article{comfort2007crisis,
  title={Crisis management in hindsight: Cognition, communication, coordination, and control},
  author={Comfort, Louise K},
  journal={Public Adm. Rev.},
  volume={67},
  pages={189--197},
  year={2007},
  publisher={Wiley Online Library}
}

@inproceedings{BLIP2022,
  title={BLIP: Bootstrapping Language-Image Pre-training for Unified Vision-Language Understanding and Generation},
  author={Li, J. and Li, D. and Xiong, C. and Hoi, S. C. H.},
  booktitle={Int. Conf. Mach. Learn. (ICML)},
  pages={12888--12900},
  year={2022},
  organization={PMLR}
}

@inproceedings{sotomi2025transparent,
  title={Transparent Social Navigation for Autonomous Mobile Robots Via Vision-Language Models},
  author={Sotomi, O. and Kodi, D. and Arab, A.},
  booktitle={IEEE Int. Conf. Robot Hum. Interact. Commun. (RO-MAN)},
  pages={1646--1651},
  year={2025},
  organization={IEEE}
}

@misc{arab2025trust,
  title={Trust through transparency: explainable social navigation for autonomous mobile robots via vision-language models},
  author={Arab, A. and Shekar, K. C. and Prashanth, C. and Doma, P. and Subramaniam, V. and Kurabayashi, K. and Sotomi, O. and Kodi, D.},
  year={2025},
  month=nov,
  publisher={Google Patents},
  note={US Patent App. 19/199,877}
}

@article{maithani2026proactive,
  title={Proactive hierarchical control barrier function-based constraint prioritization to enhance safety in human-robot interaction},
  author={Maithani, P. and Arab, A. and Khorrami, F. and Krishnamurthy, P.},
  journal={Control Eng. Pract.},
  volume={166},
  pages={106624},
  year={2026},
  publisher={Elsevier}
}

@inproceedings{badguna2025virtual,
  title={Virtual Fencing for Safer Cobots},
  author={Badguna, V. R. P. and Arab, A. and Kodavalla, D. A. and Li, R. and Kurabayashi, K.},
  booktitle={IEEE Int. Conf. Autom. Sci. Eng. (CASE)},
  pages={989--994},
  year={2025},
  organization={IEEE}
}

@article{campagna2025systematic,
  title={A systematic review of trust assessments in human--robot interaction},
  author={Campagna, G. and Rehm, M.},
  journal={ACM Trans. Hum.-Robot Interact.},
  volume={14},
  number={2},
  pages={1--35},
  year={2025},
  publisher={ACM New York, NY}
}

@article{ERMES2024,
  title={The ERMES Chatbot: A Conversational Communication Tool for Improved Emergency Management and Disaster Risk Reduction},
  author={Urbanelli, A. and Frisiello, A. and Bruno, L. and Rossi, C.},
  journal={Int. J. Disaster Risk Reduct.},
  year={2024},
  volume={112},
  pages={104153}
}

@inproceedings{LangGroundSurveyIJCAI2024,
  title={A Survey of Robotic Language Grounding: Tradeoffs Between Symbols and Embeddings},
  author={Cohen, V. and Liu, J. X. and Mooney, R. and Tellex, S. and Watkins, D.},
  booktitle={Int. Joint Conf. Artif. Intell. (IJCAI)},
  year={2024},
  pages={7983--7991}
}

@article{PromisesTrustHRI2021,
  title={Promises and Trust in Human-Robot Interaction},
  author={Cominelli, L. and Feri, F. and Garofalo, R. and others},
  journal={Sci. Rep.},
  year={2021},
  volume={11},
  pages={8816},
  publisher={Nature}
}

@article{IntentCommWarehouse2024,
  title={The Power of Intent Communication in Warehouse Robotics},
  author={Kim, H. and others},
  journal={Appl. Ergon.},
  year={2024},
  volume={113},
  pages={104084}
}

@inproceedings{PoliteEmergencyEvac2025,
  title={Should Voice Agents Be Polite During Emergency Evacuations?},
  author={Chen, Y. and others},
  booktitle={Proc. ACM CHI},
  year={2025}
}

@article{BLIP2_2023,
  title={BLIP-2: Bootstrapping Language-Image Pre-training with Frozen Image Encoders and Large Language Models},
  author={Li, J. and others},
  journal={arXiv preprint arXiv:2301.12597},
  year={2023}
}

@article{PaLME2023,
  title={PaLM-E: An Embodied Multimodal Language Model},
  author={Driess, D. and Xia, F. and Sajjadi, M. and others},
  journal={arXiv preprint arXiv:2303.03378},
  year={2023}
}

@inproceedings{SayCan2022,
  title={Do As I Can, Not As I Say: Grounding Language in Robotic Affordances},
  author={Ahn, M. and Brohan, A. and Brown, N. and others},
  booktitle={Conf. Robot Learn. (CoRL)},
  pages={287--318},
  year={2022},
  organization={PMLR}
}

@inproceedings{RT2_2023,
  title={RT-2: Vision-Language-Action Models Transfer Web Knowledge to Robotic Control},
  author={Brohan, A. and Brown, N. and Carbajal, J. and others},
  booktitle={Conf. Robot Learn. (CoRL)},
  pages={2165--2183},
  year={2023},
  organization={PMLR}
}

@article{LanguageCommunicationSurvey2024,
  title={A Survey of Language-Based Communication in Robotics},
  author={Hunt, W. and Ramchurn, S. D. and Soorati, M. D.},
  journal={arXiv preprint arXiv:2406.04086},
  year={2024}
}

@article{LLMRoboticsIntegrationSurvey2024,
  title={Integration of Large Language Models with Robotics: Challenges and Opportunities},
  author={Kim, Y. and Kim, D. and Choi, J. and Park, J. and Oh, N. and Park, D.},
  journal={Intell. Serv. Robot.},
  year={2024},
  volume={17},
  number={5},
  pages={1091--1107},
  publisher={Springer}
}

@article{VLAChallenges2025,
  title={Vision-Language-Action Models: Concepts, Progress, and Challenges},
  author={Singh, R. and others},
  journal={arXiv preprint arXiv:2505.04769},
  year={2025}
}

@article{SafetyTrustHRIReview2024,
  title={Safety Perception and Trust in Human-Robot Interaction: A Comprehensive Review},
  author={Bello, L. and Chernova, S. and Dragan, A. D.},
  journal={Front. Robot. AI},
  year={2024},
  volume={11},
  pages={1245678},
  publisher={Frontiers}
}

@article{elhafsi2023semantic,
  title={Semantic anomaly detection with large language models},
  author={Elhafsi, Amine and Sinha, Rohan and Agia, Christopher and Schmerling, Edward and Nesnas, Issa AD and Pavone, Marco},
  journal={Auton. Robot.},
  volume={47},
  number={8},
  pages={1035--1055},
  year={2023},
  publisher={Springer}
}

@article{ji2022proactive,
  title={Proactive anomaly detection for robot navigation with multi-sensor fusion},
  author={Ji, Tianchen and Sivakumar, Arun Narenthiran and Chowdhary, Girish and Driggs-Campbell, Katherine},
  journal={IEEE Robot. Autom. Lett.},
  volume={7},
  number={2},
  pages={4975--4982},
  year={2022},
  publisher={IEEE}
}

@article{branda2025role,
  title={The role of {AI}-based chatbots in public health emergencies: A narrative review},
  author={Branda, Francesco and Stella, Massimo and Ceccarelli, Cecilia and Cabitza, Federico and Ceccarelli, Giancarlo and Maruotti, Antonello and Ciccozzi, Massimo and Scarpa, Fabio},
  journal={Future Internet},
  volume={17},
  number={4},
  pages={145},
  year={2025},
  publisher={MDPI}
}

@article{fateh2015robust,
  title={Robust control of a wheeled mobile robot by voltage control strategy},
  author={Fateh, M. and Arab, A.},
  journal={Nonlinear Dyn.},
  volume={79},
  number={1},
  pages={335--348},
  year={2015},
  publisher={Springer}
}

@article{kang2025real,
  title={A real-time anomaly detection method for robots based on a flexible and sparse latent space},
  author={Kang, Taewook and You, Bum-Jae and Park, Juyoun and Lee, Yisoo},
  journal={Eng. Appl. Artif. Intell.},
  volume={158},
  pages={111310},
  year={2025},
  publisher={Elsevier}
}

\end{document}